\documentclass[12pt]{iopart}

\usepackage{iopams} 

\expandafter\let\csname equation*\endcsname\relax
\expandafter\let\csname endequation*\endcsname\relax

\usepackage{amsmath}
\usepackage{booktabs}
\usepackage{subcaption}

\usepackage[english]{babel}
\usepackage{graphicx}
\usepackage[colorlinks=true, allcolors=blue]{hyperref}
\usepackage{amssymb}
\usepackage{bm}
\usepackage{scalerel}
\usepackage[font=footnotesize]{caption}

\newcommand{\newblock}{}  %

\usepackage{xcolor}
\usepackage[normalem]{ulem}
\definecolor{tangerine}{rgb}{0.944,0.522,0}
\definecolor{verde}{rgb}{0.,0.6,0}
\definecolor{rosso}{rgb}{0.9,0.0,0.2}
\definecolor{magenta}{rgb}{0.9,0.2,0.9}

\newif\ifhighlight

\newcommand{\highlight}{\highlighttrue}
\highlight
\newcommand{\editor}[2]{%
  \expandafter\newcommand\csname #1note\endcsname[1]{%
    \textcolor{#2}{(\textbf{#1note:} \textsc{##1})}}%
  \expandafter\newcommand\csname #1\endcsname[1]{%
    \ifhighlight\textcolor{#2}{##1} \else ##1\fi}%
  \expandafter\newcommand\csname #1cancel\endcsname[1]{%
    \ifhighlight\textcolor{#2}{\sout{##1}}\fi}%
  \expandafter\newcommand\csname #1change\endcsname[2]{%
    \ifhighlight\textcolor{#2}{\sout{##1} ##2}\else ##2\fi}%
  \newenvironment{#1text}{\ifhighlight\color{#2}\fi}{\color{black}}
}

\editor{FG}{rosso}
\editor{MC}{blue}
\editor{FB}{verde}
\editor{SC}{orange}
\editor{resub}{cyan}

\begin{document}

\title[Low-cost uncertainties in trained neural networks]{A prediction rigidity formalism for low-cost uncertainties in trained neural networks}

\author{Filippo Bigi, Sanggyu Chong, Michele Ceriotti, and Federico Grasselli}

\address{Laboratory of Computational Science and Modeling, Institut des Mat\'eriaux, \'Ecole Polytechnique F\'ed\'erale de Lausanne, 1015 Lausanne, Switzerland}
\ead{\{filippo.bigi, sanggyu.chong, michele.ceriotti, federico.grasselli\}@epfl.ch}
\vspace{10pt}
\begin{indented}
\item[]February 2024
\end{indented}

\begin{abstract}
Regression methods are fundamental for scientific and technological applications. However, fitted models can be highly unreliable outside of their training domain, and hence the quantification of their uncertainty is crucial in many of their applications. Based on the solution of a constrained optimization problem, we propose ``prediction rigidities'' as a method to obtain uncertainties of arbitrary pre-trained regressors. We establish a strong connection between our framework and Bayesian inference, and we develop a last-layer approximation that allows the new method to be applied to neural networks. This extension affords cheap uncertainties without any modification to the neural network itself or its training procedure. We show the effectiveness of our method on a wide range of regression tasks, ranging from simple toy models to applications in chemistry and meteorology.
\end{abstract}

\section{Introduction}
\label{sec:intro}

Machine learning is having a large impact on many fields, from the recognition and generation of text, images and speech to applications in science, engineering and daily life tasks. In particular, deep learning~\cite{lecun2015deep, bengio2017deep} has emerged as a theoretically sound and practical approach, showing impressive flexibility and achieving scalability to huge datasets.

An accurate estimation of uncertainties in the predictions of machine learning models allows the user to assess their confidence, which is extremely useful in a wide range of contexts, as well as critical in a number of their real-world applications such as medicine, autonomous driving, weather forecasts, and autonomous laboratories~\cite{gawlikowski2023survey}. Uncertainty estimates also enable the improvement of the quality and versatility of data-driven models via active learning techniques~\cite{settles2009active}, which involve selecting highly uncertain data, adding it to the training set, and updating the model accordingly.

However, the use of uncertainty quantification schemes often faces practical challenges. Indeed, many approaches require modifications to the model or the training procedure, are very expensive, and may not easily scale to large datasets. In particular, state-of-the-art uncertainty quantification methods based on ensembles \cite{lakshminarayanan2017simple} are several times more expensive to train and evaluate than single neural networks, and cannot be used on pre-trained models. Further research on ensemble approaches is ongoing, and it has produced methods which can train and/or evaluate at competitive efficiency to single models~\cite{kunapuli2023ensemble}. Yet, these methods often deteriorate the quality of the ensemble itself~\cite{ashukha2020pitfalls}. The novel scheme in Ref.~\cite{kellner2024uncertainty} avoids such issues, but it still requires modifications to the model and its training procedure, and it cannot be applied to pre-trained networks. 

In this work, we propose a novel formalism to obtain cheap uncertainty predictions based on the solution of a constrained minimization problem, which effectively probes the ``rigidity'' of the predictions of a regression model. After a brief survey of uncertainty quantification in neural networks, we introduce the new method and establish its connection to Bayesian inference. Subsequently, we discuss its application to neural networks, which is based on the treatment of the last layer of deep learning architectures as a linear Gaussian process. Given that a vast majority of neural-network-based regression models exhibit a final linear readout layer, the proposed approach is almost universally applicable. We conclude by showing the accuracy and versatility of our method on a range of machine learning models and tasks.

\section{Background}
\label{sec:background}

\subsection{Existing uncertainty quantification schemes}

Over the years, multiple uncertainty quantification methods have been proposed for neural networks.
In this section, we will introduce a number of them, prioritizing those that are most widely used and/or those that are mathematically related to our construction. For more comprehensive reviews, we redirect the reader to Refs.~\cite{abdar2021review} and \cite{gawlikowski2023survey}.

Many attempts to provide uncertainty estimates in deep learning rely on a Bayesian formalism \cite{bernardo2009bayesian}. Such methods aim to find a posterior probability distribution over the weights, which allows one to calculate uncertainties for the output(s) of the model \cite{neal2012bayesian}. However, a full Bayesian treatment of neural networks is computationally unfeasible, and many approximate methods have been developed as a result \cite{neal2012bayesian, ritter2018scalable, malmstrom2023fusion, graves2011practical, blundell2015weight, hernandez2015probabilistic, malinin2018predictive, liu2020simple, wilson2016deep}. The most popular of these is arguably Monte Carlo dropout \cite{gal2016dropout}, which uses dropout \cite{srivastava2014dropout} at prediction time, requiring only small modifications of the underlying architecture but incurring in a substantial inference overhead (generally by a factor of 10-100).

Another Bayesian direction of research on uncertainty estimates is based on the reformulation of neural networks as Gaussian processes \cite{neal1996priors, williams1996computing, lee2017deep}. Despite their strong theoretical underpinnings, these methods are often too expensive to be of practical use, and they exhibit poor scaling to large training sets.

Several ensemble approaches have also emerged, and although these do not necessarily stem from a Bayesian formalism, much recent work has been focused on their connection to Bayesian inference \cite{he2020bayesian, hoffmann2021deep}. Most notably, deep ensembles \cite{lakshminarayanan2017simple} have shown to afford state-of-the-art uncertainty predictions on both in-domain \cite{lakshminarayanan2017simple, ashukha2020pitfalls, rahaman2021uncertainty} and out-of-domain \cite{lakshminarayanan2017simple, ovadia2019can} evaluation. In these tasks, deep ensembles are often competitive with or outperform more sophisticated methods, but they require the training and evaluation of multiple neural networks ($\geq$ 5-10). This leads to additional costs in computational time and/or resources, which can become prohibitive for sufficiently large models. Moreover, although they are relatively practical for use in classification, deep ensembles require modifications of the architecture and the training procedure if they are to be employed in regression tasks \cite{lakshminarayanan2017simple}. Since the training of deep ensembles can be very expensive, several methods to accelerate it have been proposed \cite{huang2017snapshot, garipov2018loss, zhang2019cyclical}. However, these have been shown to induce additional correlations among the individual members of the ensembles, so that it is then necessary to employ larger ensembles to recover the same uncertainty prediction quality as standard deep ensembles \cite{ashukha2020pitfalls}, which results in an even larger prediction overhead. As a result, Ref.~\cite{kellner2024uncertainty} proposes a weight-sharing solution that affords good uncertainty estimates with virtually no training or prediction overhead. However, this and other similar approaches still require modifications to the original model architecture, which is not always straightforward or practical, and cannot provide uncertainty estimates for models that have already been trained.

\subsection{Related work}

Few methods in the literature are particularly relevant to our theoretical approach, and we will discuss them in more detail here.
\begin{itemize}
    \item \textbf{Laplace approximation} \cite{laplace1774memoire}. This is one of the oldest Bayesian method to be applied to neural networks. In this framework, the weights are considered to be stochastic variables, and the loss function is identified as the negative logarithm of a likelihood function that aims to be maximized \cite{daxberger2021laplace}. The Laplace approximation consists in approximating the resulting weight probability distribution as a multivariate Gaussian. Uncertainty estimates on the targets are then obtained via first-order uncertainty propagation.
    \item \textbf{Deep kernel learning} \cite{wilson2016deep}. This method learns kernel functions via a deep architecture. Based on the KISS-GP kernel approximation \cite{wilson2015kernel}, it achieves $\mathcal{O}(1)$-scaling predictions, as opposed to traditional Gaussian process regressors, whose target and uncertainty predictions scale as $\mathcal{O}(N_{\mathrm{train}})$ and $\mathcal{O}(N_{\mathrm{train}}^2)$, respectively. However, this method requires a specific architecture and training process, and its uncertainty estimates have been shown to exhibit pathological behavior even in some simple cases \cite{ober2021promises}.
    \item \textbf{Neural networks as Gaussian processes}. The concepts of Neural Network Gaussian Process (NNGP, \cite{lee2017deep}) and Neural Tangent Kernel (NTK, \cite{jacot2018neural}) have been developed as an effort to understand deep learning and neural networks from a theoretical perspective. NNGPs can be seen as the Gaussian process equivalents of arbitrary neural networks architectures in the infinite-width limit, while NTKs are fundamental in describing their linearized training dynamics.
    
\end{itemize}

\section{Theory}
\label{sec:theory}

\subsection{Problem statement and notation}

We consider a regression task where the training set $\mathcal{D}$ is composed of pairs $\{\mathbf{x}_i, y_i\}_{i=1}^{N_{\mathrm{train}}}$, where $\mathbf{x}_i \in \mathbb{R}^d$ are the input features for each training sample $i$, and $y_i \in \mathbb{R}$ are the corresponding targets. Although the choice of single targets might seem restrictive, it would be relatively straightforward to generalize the results to multiple regression targets. The predictions $\tilde{y}$ of the regression model are given by $\tilde{y}_i \equiv \tilde{y}(\mathbf{x}_i, \mathbf{w})$. The loss function associated with the task is defined as a sum over individual contributions in the training set:

\begin{equation}
    \mathcal{L}(\mathbf{w}) = \sum_{i=1}^{N_{\rm{train}}} \ell_i \equiv \sum_{i=1}^{N_{\rm{train}}} \ell(\tilde{y}_i, y_i) \equiv \sum_{i=1}^{N_{\rm{train}}} \ell(\tilde{y}(\mathbf{x}_i, \mathbf{w}), y_i).
\end{equation}

\subsection{Prediction rigidities as the solution of a constrained optimization problem}

The notion of prediction rigidity was originally conceived in the field of atomistic modeling, where the total energy of a chemical structure is predicted as the sum of local contributions \cite{chong2023robustness}. Although such local energies are not physical observables \textit{per se}, local predictions have been used widely in the computational chemistry community, due to their strong heuristic power in describing the local energetics of chemophysical processes. In this context, local prediction rigidities constitute a measure of how well-defined local energies are. In this section, we re-derive part of the results to show that the same concepts developed in Ref.~\cite{chong2023robustness} can be borrowed to quantify uncertainties on generic targets.

To approach the robustness of a prediction on a single sample (labeled as $\star$), we can consider how sensitive the model is to a change in the prediction of that sample. To do so, a new loss $\mathcal{L}_c$ can be defined to include a Lagrangian term that constrains the prediction for $\star$, i.e. $\tilde{y}(\mathbf{x}_\star, \mathbf{w})$, to an arbitrary value $\epsilon_\star$:
\begin{equation}
    \mathcal{L}_c(\mathbf{w}, \lambda, \epsilon_\star) = \mathcal{L}(\mathbf{w}) + \lambda (\epsilon_\star - \tilde{y}(\mathbf{x}_\star, \mathbf{w})).
\end{equation}
After solving for $\partial \mathcal{L}_c / \partial \mathbf{w} = \mathbf{0}$ and $\partial \mathcal{L}_c / \partial \lambda = 0$ to find the minimum of the constrained model, the rigidity of the prediction $\tilde{y}(\mathbf{x}_\star, \mathbf{w})$ can be defined as $R_\star = \partial \mathcal{L}_c(\epsilon_\star) / \partial \epsilon_\star^2 |_{\epsilon_\star = \tilde{y}(\mathbf{x}_\star, \mathbf{w}_o)} $. Intuitively, predictions which are less rigid with respect to a perturbation will be less confident, and vice versa. The connection between prediction rigidities and Bayesian inference will be discussed in Sec. \ref{sec:pr-bayesian}. 

Although the equations of the constrained minimization problem are difficult to solve in general, a few common approximations allow them to be solved analytically.
We perform a second-order expansion of the unconstrained loss function $\mathcal{L}$ around the optimal weights $\mathbf{w}_o$:
\begin{equation}\label{eq:loss-2nd-order}
    \mathcal{L}(\mathbf{w}) \approx \mathcal{L}(\mathbf{w}_o) +\frac{1}{2} (\mathbf{w}-\mathbf{w}_o)^\top \mathbf{H}_o (\mathbf{w}-\mathbf{w}_o),
\end{equation}
where the first-order term vanishes due to the optimality condition and
\begin{equation}
    \mathbf{H}_o = \frac{\partial^2 \mathcal{L}}{\partial \mathbf{w} \, \partial \mathbf{w}^\top} \Big|_{\mathbf{w}_o}
\end{equation}
is the Hessian at the optimum. Now, we will solve $\partial \mathcal{L}_c / \partial \mathbf{w} = \mathbf{0}$ and $\partial \mathcal{L}_c / \partial \lambda = 0$, naming $\mathbf{w}_c$ and $\lambda_c$ as the solutions to the constrained minimization problem:

\begin{equation}\label{eq:approx_L_c}
    \mathcal{L}_c(\mathbf{w}, \lambda, \epsilon_\star)
    \stackrel{\eqref{eq:loss-2nd-order}}{\approx} \mathcal{L}(\mathbf{w}_o) + \frac{1}{2} (\mathbf{w}-\mathbf{w}_o)^\top \mathbf{H}_o (\mathbf{w}-\mathbf{w}_o) + \lambda (\epsilon_\star - \tilde{y}(\mathbf{x}_\star, \mathbf{w}))
\end{equation}

\begin{equation}\label{eq:der_w-w0_constr_eq}
    \frac{\partial \mathcal{L}_c}{\partial \mathbf{w}} \Big|_{\mathbf{w}_c, \lambda_c} = \mathbf{H}_o (\mathbf{w}_c-\mathbf{w}_o) - \lambda_c \frac{\partial \tilde{y}_\star}{\partial \mathbf{w}} \Big|_{\mathbf{w}_c} = \mathbf{0} \,\, \Longrightarrow \,\, \mathbf{w}_c-\mathbf{w}_o = \lambda_c \, \mathbf{H}_o^{-1} \frac{\partial \tilde{y}_\star}{\partial \mathbf{w}} \Big|_{\mathbf{w}_c}
\end{equation}
\begin{equation}
   \frac{\partial \mathcal{L}_c}{\partial \lambda} \Big|_{\mathbf{w}_c, \lambda_c} = \epsilon_\star - \tilde{y}(\mathbf{x}_\star, \mathbf{w}_c) = 0
   \label{eq:der_lambda_constr_eq}
\end{equation}
Substituting Eqs.~\eqref{eq:der_w-w0_constr_eq} and~\eqref{eq:der_lambda_constr_eq} into Eq.~\eqref{eq:approx_L_c}, we obtain
\begin{equation}
    \mathcal{L}_c(\mathbf{w}_c, \lambda_c, \epsilon_\star) 
    \stackrel{\eqref{eq:loss-2nd-order}}{\approx} \mathcal{L}(\mathbf{w}_o) + \frac{1}{2} \lambda_c^2 \, \frac{\partial \tilde{y}_\star}{\partial\mathbf{w}} \Big|^\top_{\mathbf{w}_o} \mathbf{H}_o^{-1} \frac{\partial \tilde{y}_\star}{\partial \mathbf{w}} \Big|_{\mathbf{w}_c}
\end{equation}
From a linearization of the predictions near the optimum of the unconstrained loss in the latter expression, we obtain:
\begin{equation}\label{eq:equal-gradients}
\frac{\partial \tilde{y}_\star}{\partial \mathbf{w}} \Big|_{\mathbf{w}_c} = \frac{\partial \tilde{y}_\star}{\partial \mathbf{w}} \Big|_{\mathbf{w}_o}
\end{equation}
and
\begin{equation}\label{eq:lin_constr}
    \epsilon_\star - \tilde{y}(\mathbf{x}_\star, \mathbf{w}_c) \approx \epsilon_\star - \tilde{y}(\mathbf{x}_\star, \mathbf{w}_o) - \frac{\partial \tilde{y}_\star}{\partial\mathbf{w}} \Big|^\top_{\mathbf{w}_o} (\mathbf{w}_c-\mathbf{w}_o) = 0.
\end{equation}
Substituting Eq.~\eqref{eq:der_w-w0_constr_eq} into Eq.~\eqref{eq:lin_constr} yields
\begin{equation}
    \epsilon_\star - \tilde{y}(\mathbf{x}_\star, \mathbf{w}_o) - \lambda_c \frac{\partial \tilde{y}_\star}{\partial\mathbf{w}} \Big|^\top_{\mathbf{w}_o} \mathbf{H}_o^{-1} \frac{\partial \tilde{y}_\star}{\partial \mathbf{w}} \Big|_{\mathbf{w}_c} = 0 \,\, \Longrightarrow \,\, \lambda_c = \frac{\epsilon_\star - \tilde{y}(\mathbf{x}_\star, \mathbf{w}_o)}{\frac{\partial \tilde{y}_\star}{\partial\mathbf{w}} \Big|^\top_{\mathbf{w}_o} \mathbf{H}_o^{-1} \frac{\partial \tilde{y}_\star}{\partial \mathbf{w}} \Big|_{\mathbf{w}_c}},
    \label{eq:lambda_c_final}
\end{equation}
and, finally, using Eqs.~\eqref{eq:equal-gradients} and~\eqref{eq:lambda_c_final} into Eq.~\eqref{eq:approx_L_c} affords:
\begin{equation}\label{eq:loss-2nd-order-with-lambda}
    \mathcal{L}_c(\epsilon_\star) \approx \mathcal{L}(\mathbf{w}_o) + \frac{1}{2} \, \frac{(\epsilon_\star - \tilde{y}(\mathbf{x}_\star, \mathbf{w}_o))^2}{\frac{\partial \tilde{y}_\star}{\partial\mathbf{w}} \Big|^\top_{\mathbf{w}_o} \mathbf{H}_o^{-1} \frac{\partial \tilde{y}_\star}{\partial \mathbf{w}} \Big|_{\mathbf{w}_o}}.
\end{equation}
The prediction rigidity $R_\star$ can then be found as
\begin{equation}
    R_\star \equiv \frac{\partial^2 \mathcal{L}_c(\epsilon_\star)}{\partial \epsilon_\star^2} \Big|_{\epsilon_\star = \tilde{y}(\mathbf{x}_\star, \mathbf{w}_o)} = \Big( \frac{\partial \tilde{y}_\star}{\partial\mathbf{w}} \Big|^\top_{\mathbf{w}_o} \mathbf{H}_o^{-1} \frac{\partial \tilde{y}_\star}{\partial \mathbf{w}} \Big|_{\mathbf{w}_o} \Big)^{-1}.\label{eq:PR}
\end{equation}

\subsection{Prediction rigidities and Bayesian inference}\label{sec:pr-bayesian}

To make the connection between prediction rigidities and Bayesian inference, we begin by interpreting the constraint as a stochastic variable, $\epsilon_\star \rightarrow \hat{\epsilon}_\star$, whose maximum a posteriori (MAP) estimate is the solution to the unconstrained problem, $\tilde{y}(\mathbf{x}_\star, \mathbf{w}_o)$. The constrained loss then becomes a function of a stochastic variable $\hat{\epsilon}_\star$, i.e.~$\mathcal{L}_c = \mathcal{L}_c(\hat{\epsilon}_\star|\mathbf{x}_\star,\mathcal{D})$. We then identify $\mathcal{L}_c$ as the negative logarithm of an unnormalized probability distribution $Z p(\hat{\epsilon}_\star|\mathbf{x}_\star, \mathcal{D})$, where $p$ is a normalized distribution of the output, given the input $\mathbf{x}_\star$ and the training dataset $\mathcal{D}$, while $Z$ is a normalization constant. This is a standard operation whenever a likelihood-based loss function is assumed \cite{daxberger2021laplace,hastie2009elements}. 
Hence, our constrained minimization formulation, based on a second-order Taylor expansion of the loss, is in practice equivalent to a Laplace approximation of the probability distribution of the outputs, $p(\hat{\epsilon}_\star|\mathbf{x}_\star, \mathcal{D})$:
\begin{equation}
    p(\hat{\epsilon}_\star|\mathbf{x}_\star, \mathcal{D}) \approx
    \mathcal{N}(\tilde{y}(\mathbf{x}_\star, \mathbf{w}_o), R_\star^{-1})
\end{equation}
Therefore, in this probabilistic picture, we can identify the curvature of the loss in Eq.~\eqref{eq:PR} as the reciprocal of the variance associated to the prediction $\tilde{y}(\mathbf{x}_\star,\mathbf{w}_o)$, i.e.~its uncertainty.

\subsection{An efficient approximation for the Hessian}

It should be noted that the Hessian of the loss at the optimum, $\mathbf{H}_o$, can be pre-computed based only on the training set, and therefore it does not affect inference time for the evaluation of the uncertainties. Despite this fact, the exact Hessian of the loss can be exceedingly expensive to calculate. Therefore, we approximate $\mathbf{H}_o$ as
\begin{multline}
    \mathbf{H}_o = \frac{\partial^2 \mathcal{L}}{\partial \mathbf{w} \, \partial \mathbf{w}^\top} = \frac{\partial}{\partial \mathbf{w}} \sum_i \frac{\partial \ell_i}{\partial \mathbf{w}^\top} = \frac{\partial}{\partial \mathbf{w}} \sum_i \frac{\partial \ell_i}{\partial \tilde{y}_i} \frac{\partial \tilde{y}_i}{\partial \mathbf{w}^\top} = \sum_i \Big( \frac{\partial \ell_i}{\partial \tilde{y}_i} \frac{\partial^2 \tilde{y}_i}{\partial \mathbf{w} \partial \mathbf{w}^\top} + \frac{\partial}{\partial \mathbf{w}} \frac{\partial \ell_i}{\partial \tilde{y}_i} \frac{\partial \tilde{y}_i}{\partial \mathbf{w}^\top} \Big) = \\ \sum_i \frac{\partial \ell_i}{\partial \tilde{y}_i} \frac{\partial^2 \tilde{y}_i}{\partial \mathbf{w} \partial \mathbf{w}^\top} + \sum_i \frac{\partial \tilde{y}_i}{\partial \mathbf{w}} \frac{\partial^2 \ell_i}{\partial \tilde{y}_i^2} \frac{\partial \tilde{y}_i}{\partial \mathbf{w}^\top} \approx \sum_i \frac{\partial \tilde{y}_i}{\partial \mathbf{w}} \frac{\partial^2 \ell_i}{\partial \tilde{y}_i^2} \frac{\partial \tilde{y}_i}{\partial \mathbf{w}^\top},
\end{multline}
where we take each derivative at the optimum (although we do not write it explicitly for simplicity of notation). The final equality assumes the first term in the penultimate expression to be negligible at the optimum. This approximation is not only reasonable because $\partial \tilde{y}_i / \partial \mathbf{w}$ is close to zero in accurate models, but it is also advantageous in case of outliers and it has therefore in use in several successful optimization algorithms \cite{levenberg1944method, marquardt1963algorithm}. For a more thorough justification of this approximation, we redirect the reader to Ref.~\cite{press2007numerical}, Sec.~15.5.2. Most importantly, the resulting pseudo-Hessian matrix can be calculated without the need for second derivatives of the model $\tilde{y}$. 

It should be noted that, even within this approximation, prediction rigidities provide the correct analytical uncertainty estimates for linear regression as well as Gaussian process regression. This is shown in~\ref{app:linear-gpr}, and it is particularly important for the application of the approximation to neural networks.

\subsection{Application to neural networks}\label{sec:nn-application}
\newcommand{\nngp}{{\scaleto{\mathrm{NNGP}}{4pt}}}
\newcommand{\ntk}{{\scaleto{\mathrm{NTK}}{4pt}}}

We now extend the prediction rigidity formalism to neural network models. In particular, we consider a trained neural network with a final readout layer which operates on $N_L$ latent features $\{\mathbf{f}_i\}_{i=1}^{N_L}$ and predicts a single regression target. In other words, this layer applies a linear transformation from $\mathbb{R}^{N_L}$ to $\mathbb{R}$. We will often make use of a matrix $\mathbf{F}$ (size $N_{\rm{train}} \times N_L$) which collects all last-layer latent features in the training set $\mathcal{D}$, as well as a vector $\mathbf{y}$ that collects all targets $y_i$ in the training set $\mathcal{D}$. Although we do not explicitly consider any bias terms in the following discussion, we discuss their influence in~\ref{app:biases}.

It is clear that prediction rigidities cannot scale to large neural networks due to the quadratic memory requirements to store $\mathbf{H}_o$. Therefore, we propose a simple but effective last-layer approximation, which identifies the PR as the reciprocal of
\begin{equation}\label{eq:llpr}
    \sigma_\star^2 \propto \frac{\partial \tilde{y}_\star}{\partial\mathbf{w}} \Big|^\top_{\mathbf{w}_o} \mathbf{H}_o^{-1} \frac{\partial \tilde{y}_\star}{\partial \mathbf{w}} \Big|_{\mathbf{w}_o} \approx  \mathbf{f}_\star^\top (\mathbf{F}^\top \mathbf{F})^{-1} \mathbf{f}_\star, 
\end{equation}
where the last equality assumes that all layers before the final linear layer do not contribute to the uncertainty of the predictions. We now proceed to justify this approximation in terms of the equivalence between wide neural networks and Gaussian processes.

By showing that the functional form and initialization scheme of infinitely wide networks directly implies a kernel (covariance) function, Ref.~\cite{lee2017deep} effectively proved the equivalence between deep neural networks and Gaussian processes. The corresponding ``neural network Gaussian processes'' (NNGPs) can be trained and predict as pure Gaussian processes, and they often outperform the corresponding finite-width neural networks.

More recently, Ref.~\cite{lee2019wide} showed that wide neural networks behave as Gaussian processes also during their training process. In particular, they showed that, for a learning rate $\eta < \eta_{\mathrm{max}}$, they evolve as their linearized counterpart under gradient descent, meaning that their neural tangent kernel \cite{jacot2018neural}, defined as
\begin{equation}\label{eq:def_NTK}
    K_\ntk(\mathbf{x}_i, \mathbf{x}_j) = \Big( \frac{\partial \tilde{y}(\mathbf{x}_i, {\mathbf{w}})}{\partial \mathbf{w}} \Big)^\top \, \frac{\partial \tilde{y}(\mathbf{x}_j, {\mathbf{w}})}{\partial \mathbf{w}},
\end{equation}
remains arbitrarily close (as the width goes to infinity) to its initial value during the entire learning trajectory. Here, ${\mathbf{w}}$ are the free parameters of the neural network (weights and biases) and $\tilde{y}$ is the neural network function. In addition, they show that, under these assumptions and for simple regression tasks with a squared-error-type loss, the Gaussian process representing the neural network during training has mean
\begin{equation}
    \mu_\star = \mathbf{k}_\ntk(\star, \mathcal{D}) \, \mathbf{K}^{-1}_\ntk \, (\mathbf{I} - e^{-\eta \mathbf{K}_\ntk t})  \,\mathbf{y}
\end{equation}
and variance 
\begin{multline}\label{eq:nngp-ntk-variance}
    \sigma_\star^2 = k_\nngp(\star, \star) \\ 
    + \mathbf{k}_\ntk(\star, \mathcal{D}) \, \mathbf{K}^{-1}_\ntk \, (\mathbf{I} - e^{-\eta \mathbf{K}_\ntk t}) \, \mathbf{K}_\nngp \, (\mathbf{I} - e^{-\eta \mathbf{K}_\ntk t}) \, \mathbf{K}^{-1}_\ntk \, \mathbf{k}_\ntk(\mathcal{D}, \star)
    \\ 
    - \mathbf{k}_\ntk(\star, \mathcal{D}) \, \mathbf{K}^{-1}_\ntk \, (\mathbf{I} - e^{-\eta \mathbf{K}_\ntk t}) \, \mathbf{k}_\nngp(\mathcal{D}, \star) \\
    - \mathbf{k}_\nngp(\star, \mathcal{D}) \, (\mathbf{I} - e^{-\eta \mathbf{K}_\ntk t}) \, \mathbf{K}^{-1}_\ntk \, \mathbf{k}_\ntk(\mathcal{D}, \star).
\end{multline}
Here, $t$ is the time since the beginning of (exact) gradient descent, $\mathbf{K}_\mathrm{NTK}$ is the $N_\mathrm{train}\times N_\mathrm{train}$ matrix collecting the NTK of Eq.~\eqref{eq:def_NTK}, $\mathbf{I}$ is the identity of same size, $\mathbf{k}_\ntk(\mathcal{D}, \star)$ the vector of NTK between the training set and a test point $\star$, and $\mathbf{k}_\ntk(\star, \mathcal{D}) = \mathbf{k}_\ntk(\mathcal{D}, \star)^\top$. Analogous notation holds for NNGP kernels. 

Our simplified prediction rigidities for neural networks are based on the calculation of a simple approximation to the covariance function of the NTK. Following Ref.~\cite{lee2017deep}, if $\sigma_w^2$ are the variances of the weights of the last layer at initialization, then $K_\nngp$ is given by
\begin{equation}\label{eq:K-nngp}
    K_\nngp(\mathbf{x}_i, \mathbf{x}_j) = N_L \sigma_w^2 \, \mathbb{E}_{z^{(L-1)} \sim \mathcal{N}(0, K_\nngp^{(L-1)})}[\phi(z^{(L-1)}(\mathbf{x}_i)) \phi(z^{(L-1)}(\mathbf{x}_j))],
\end{equation}
where $z^{(L-1)}$ are the pre-activations of the last layer, which are sampled from a Gaussian process with zero mean and covariance $K_\nngp^{(L-1)}$, which is the $K_\nngp$ of the same neural network excluding the last layer. Here, we propose taking the expectation value above with respect to the different features of the last layer (which are identically distributed at initialization), so that
\begin{equation}\label{eq:nngp-approx}
    K_\nngp(\mathbf{x}_i, \mathbf{x}_j) \approx \sigma_w^2 \, \mathbf{f}_i^\top  \mathbf{f}_j \Longrightarrow \mathbf{K}_\nngp(\mathcal{D}, \mathcal{D}) \approx \sigma_w^2 \, \mathbf{F} \mathbf{F}^\top.
\end{equation}
This expression is exact in the limit of an infinite number of features in the last layer, and its error decreases as $N_L^{-1/2}$. It is important to note that, at initialization, the above expression does not rely on a last-layer approximation.

Furthermore, we compute an approximation to the NTK which only involves the weights of the last layer $\mathbf{w}_L$:
\begin{equation}\label{eq:NTK_multiple_FF^T}
    K_\ntk(\mathbf{x}_i, \mathbf{x}_j) \approx c \, \Big( \frac{\partial \tilde{y}(\mathbf{x}_i, \mathbf{w})}{\partial \mathbf{w}_L} \Big)^\top \,\, \frac{\partial \tilde{y}(\mathbf{x}_j, \mathbf{w})}{\partial \mathbf{w}_L} = c \, \mathbf{f}_i^\top  \mathbf{f}_j \Longrightarrow \mathbf{K}_\ntk(\mathcal{D}, \mathcal{D}) \approx c \, \mathbf{F} \mathbf{F}^\top,
\end{equation}
where $c$ is a constant that only depends on the architecture of the neural network. In particular, $c$ is independent of $i$ and $j$. Such an approximation becomes an equality in the case of linear activation function, and is still robust for all activation functions $\phi(z)$ which can be expanded around $z=0$, as shown in \ref{app:NTK linearization}. 

Given these approximations and $t \rightarrow +\infty$, Eq.~\eqref{eq:nngp-ntk-variance} simply reduces to
\begin{equation}
    \sigma_\star^2 = \sigma_w^2 \, (\mathbf{f}_\star^\top \mathbf{f}_\star - \mathbf{f}_\star^\top \mathbf{F}^\top (\mathbf{F} \mathbf{F}^\top)^{-1} \mathbf{F} \, \mathbf{f}_\star),
\end{equation}
which is independent of the constant $c$ introduced in Eq.~\eqref{eq:NTK_multiple_FF^T}.
We now introduce a small regularization term $\varsigma^2$, whose role is explained in~\ref{app:regularizer}, to obtain
\begin{equation}\label{eq:woodbury-trick}
    \sigma_\star^2 = \sigma_w^2 \, (\mathbf{f}_\star^\top \mathbf{f}_\star - \mathbf{f}_\star^\top \mathbf{F}^\top (\mathbf{F} \mathbf{F}^\top + \varsigma^2 \mathbf{I})^{-1} \mathbf{F} \, \mathbf{f}_\star) = \sigma_w^2 \varsigma^2 \, \mathbf{f}_\star^\top (\mathbf{F}^\top \mathbf{F} + \varsigma^2 \mathbf{I})^{-1} \mathbf{f}_\star.
\end{equation}
where the second equality makes use of the Woodbury identity~\cite{woodbury1950inverting}. The final uncertainty expression is simply
\begin{equation}\label{eq:nngp-ntk-final-variance}
    \sigma_\star^2 = \alpha^2 \, \mathbf{f}_\star^\top (\mathbf{F}^\top \mathbf{F} + \varsigma^2 \mathbf{I})^{-1} \mathbf{f}_\star.
\end{equation}
In practical applications, the product $\alpha^2 \equiv \sigma_w^2 \varsigma^2$ must be calibrated (in the same way that the analogous constant must be calibrated in linear or Gaussian process regression). Crucially, $\alpha^2$ does not depend on the specific point $\star$.

Since Eq.~\eqref{eq:nngp-ntk-final-variance} corresponds to Eq.~\eqref{eq:llpr}, we have justified the use of the last-layer PR (LLPR) formalism for NNs in terms of the linearized training dynamics of neural networks. Given the approximations involved, we expect the LLPR to better quantify model uncertainty when the width of the neural network (and, in particular, of its last layer) is large. We confirm this supposition empirically in~\ref{app:width}.

$\mathbf{F}^\top \mathbf{F}$ can be pre-computed after training in a batched manner, which avoids storing any potentially huge matrices (such as $\mathbf{F}$ itself). This step requires a single run through the training set, with a negligible cost compared to training itself, which usually consists of hundreds or thousands of passes through the training set, each of these including the calculation of gradients.

Given $\mathbf{F}^\top \mathbf{F}$, the proposed method is able to generate uncertainty estimates through a single forward pass of the neural network (the same that is used for inference), which immediately gives access to $\mathbf{f_\star}$. The only additional operation is the computation of $\mathbf{f}_\star^\top (\mathbf{F}^\top \mathbf{F} + \varsigma^2 \mathbf{I})^{-1} \mathbf{f}_\star$, where the middle term is pre-computed. This computation has the same cost as an extra layer, and it is therefore reasonable to conclude that the proposed uncertainty estimation method generates little computational overhead in the vast majority of cases. If one wanted to further decrease the cost of inference, it would be possible to pre-compute a truncated eigenvalue decomposition of $(\mathbf{F}^\top \mathbf{F} + \varsigma^2 \mathbf{I})^{-1}$, or to generate an ensemble of last-layer weights (see Sec.~\ref{sec:propagation}), which is equivalent to performing a Gaussian integral analytically.

\newcommand{\gtick}{\textcolor{green}{\checkmark}}
\newcommand{\rcross}{\textcolor{red}{$\times$}}

\begin{table}[h]
\begin{center}
\begin{tabular}{lcccc}
\toprule
Method & No UQ & MCD & DE & LLPR \\
\midrule
Training cost & 1 & $\approx 1$ & N & $\approx 1$ \\
Inference cost & 1 & M & N & $\approx 1$ \\
\bottomrule
\end{tabular}
\end{center}
\caption{
Comparison of the theoretical training and inference costs for Monte-Carlo dropout, deep ensembles and LLPR.
All costs are normalized to those of a simple model without uncertainty quantification (``No UQ''). For Monte-Carlo dropout, although Ref.~\cite{gal2016dropout} uses M\,=\,10000, more typical values are M\,=\,10-100. For deep ensembles, typical values are N\,=\,5-30 \cite{lakshminarayanan2017simple, ashukha2020pitfalls, rahaman2021uncertainty}.
}
\label{tab:comparison}
\end{table}

\subsection{Uncertainty propagation}\label{sec:propagation}

The prediction rigidity framework can also easily accommodate for uncertainty propagation, either analytically or numerically. For example, if the inputs of the machine learning model $\mathbf{x}$ are uncertain, a simple Laplace approximation in the distribution of the inputs allows to propagate the uncertainties in $\mathbf{x}$ to uncertainties of the output(s), at the cost of computing derivatives of the output(s) of the model with respect to its inputs. Assuming that inputs and weights of the model are uncorrelated, the predicted variance due to the inputs and that due to the model can simply be added.

Another scenario is that the outputs of the model are themselves needed for workflows of arbitrary complexity that generate derived quantities, i.e., functions of the regressor's output. In simple cases, this propagation of uncertainties can be performed analytically; if this is not possible, it can be achieved by generating an ensemble of weights. For example, in the neural network case, this corresponds to sampling the distribution
\begin{equation}
    \mathbf{w} \sim \mathcal{N} \big(\mathbf{w}_o, \alpha^2 (\mathbf{F}^\top \mathbf{F} + \varsigma^2 \mathbf{I})^{-1} \big)
\end{equation}
and using it to generate a last-layer ensemble, for which uncertainty propagation can be performed numerically as in Ref.~\cite{kellner2024uncertainty}. This approach is able to capture correlations between predictions, determine errors on derivatives, and in general provide accurate uncertainty predictions on any derived quantity~\cite{imbalzano2021uncertainty}.

\section{Results}
\label{sec:results}

Having described the theoretical setup and the concrete advantages of the prediction rigidity framework and its last-layer approximation, we now test the quality of its uncertainty estimates for a wide range of regression tasks.

\subsection{A simple 1D example}

\begin{figure}[h]
  \centering
  \includegraphics[width=\textwidth]{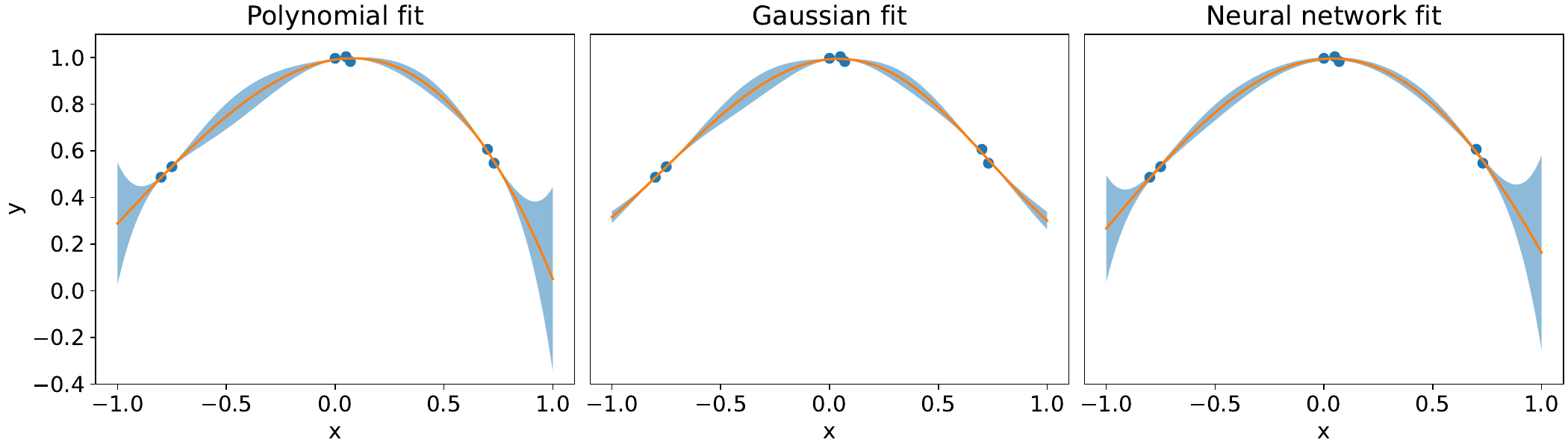}
  \caption{Uncertainties predicted as the inverse of the prediction rigidity for polynomial fit, Gaussian fit, and neural network fit (last-layer approximation), respectively. In all three cases, Training set points are marked in blue, model prediction is shown in orange, and the estimated uncertainties are shaded in light blue.}
  \label{fig:toy_fits}
\end{figure}

As a visual illustration, we consider the fit of a simple $\mathbb{R} \rightarrow \mathbb{R}$ function. We chose the $y = \mathrm{cos}^2 (x)$ function, from which we sampled 7 points and added random Gaussian noise with a standard deviation of $0.01$. We used these points as the training set to fit three different regressors: (a) a polynomial of degree 3, (b) a sum of two Gaussian functions, each containing mean and variance as  optimizable prefactors, and (c) a simple feed-forward neural network with 2 hidden layers and 32 neurons per hidden layer. Note that for the last case, the LLPR approach was adopted. The results in Figure~\ref{fig:toy_fits} reveal that the inverse of prediction rigidities provide reasonable uncertainty estimates in all three cases, reflecting the nature of the chosen regressor.

\subsection{Probabilistic backpropagation benchmark}

Next, we consider the efficacy of LLPR uncertainty quantification approach compared to established uncertainty quantification schemes for a set of regression benchmarks that were originally considered in Ref.~\cite{hernandez2015probabilistic}. Here, we concurrently consider two metrics, the first of which is the root-mean-square error $\mathrm{RMSE} = \sqrt{\frac{1}{N_{\mathrm{test}}}\sum_{t \in \mathcal{T}} (y_t - \tilde{y}_t)^2}$ (where $\mathcal{T}$ is a test set and $N_\mathrm{test}$ its size), which measures the accuracy of the network's raw predictions. The second is the negative log likelihood $\mathrm{NLL} = \frac{1}{N_{\mathrm{test}}}\sum_{t \in \mathcal{T}} \frac{1}{2}((\frac{y_t - \mu_t}{\sigma_t})^2 + \log \sigma_t^2 + \log 2\pi)$, which measures how well the predicted Gaussian distribution explains the test targets. Model training details can be found in ~\ref{app:details}.

\newcommand{\B}[1]{\textbf{#1}}  %
\newcommand{\err}[1]{{\scaleto{\mathrm{\pm#1}}{3.3pt}}}  %

\begin{table}[h]
  \centering
  \begin{tabular}{l|cccc|cccc}
  \toprule
     & \multicolumn{4}{c|}{RMSE} & \multicolumn{4}{c}{NLL} \\
    Dataset & PBP & MCD & DE & LLPR & PBP & MCD & DE & LLPR \\
  \midrule
    Concrete
    & 5.67\err{0.09} & \B{5.23}\err{0.12} & 6.03\err{0.13} & \B{5.26}\err{0.25}
    & 3.16\err{0.02} & \B{3.04}\err{0.02} & \B{3.06}\err{0.04} & \B{3.09}\err{0.07} \\
    Energy
    & 1.80\err{0.05} & 1.66\err{0.04} & 2.09\err{0.06} & \B{0.49}\err{0.03}
    & 2.04\err{0.02} & 1.99\err{0.02} & 1.38\err{0.05} & \B{0.69}\err{0.07} \\
    Kin8nm
    & 0.10\err{0.00} & 0.10\err{0.00} & 0.09\err{0.00} & \B{0.08}\err{0.00}
    & -0.90\err{0.01} & -0.95\err{0.01} & \B{-1.20}\err{0.00} & -1.12\err{0.01} \\
    Naval
    & 0.01\err{0.00} & 0.01\err{0.00} & \B{0.00}\err{0.00} & \B{0.00}\err{0.00}
    & -3.73\err{0.01} & -3.80\err{0.01} & -5.63\err{0.01} & \B{-7.07}\err{0.08} \\
    Power
    & 4.12\err{0.03} & \B{4.02}\err{0.04} & 4.11\err{0.04} & \B{3.94}\err{0.07}
    & 2.84\err{0.01} & \B{2.80}\err{0.01} & \B{2.79}\err{0.01} & 2.83\err{0.02} \\
    Protein
    & 4.73\err{0.01} & 4.36\err{0.02} & 4.71\err{0.03} & \B{4.18}\err{0.02}
    & 2.97\err{0.00} & 2.89\err{0.00} & \B{2.83}\err{0.01} & 2.91\err{0.01} \\
    Wine
    & \B{0.64}\err{0.01} & \B{0.62}\err{0.01} & \B{0.64}\err{0.01} & \B{0.63}\err{0.02}
    & 0.97\err{0.01} & \B{0.93}\err{0.01} & \B{0.94}\err{0.03} & 1.02\err{0.03} \\
    Yacht
    & \B{1.02}\err{0.05} & \B{1.11}\err{0.08} & 1.58\err{0.11} & \B{1.19}\err{0.16}
    & 1.63\err{0.02} & 1.55\err{0.03} & \B{1.18}\err{0.05} & 1.58\err{0.20} \\
    Year
    & 8.88\err{N/A} & \B{8.86}\err{N/A} & 8.89\err{N/A} & 8.91\err{N/A}
    & 3.60\err{N/A} & 3.59\err{N/A} & \B{3.35}\err{N/A} & 3.61\err{N/A} \\
  \bottomrule
  \end{tabular}
  \caption{Comparison of the performance of LLPR against probabilistic backpropagation (PBP), Monte-Carlo dropout (MCD), and deep ensembles (DE) on the benchmark datasets from Ref.~\cite{hernandez2015probabilistic}. Subscripts indicate the standard errors on 20 random splits, except for the protein dataset and the year dataset, for which 5 and 1 splits were used, respectively.}
  \label{tab:vi-benchmark}
\end{table}

Table~\ref{tab:vi-benchmark} shows that the LLPR method performs comparably with other well-established uncertainty quantification methods in terms of both RMSE and NLL. In many cases, LLPR seems to yield particularly good RMSEs. Although precise implementation details may play a role in this, it is nonetheless plausible to attribute this observation to the fact that the LLPR method simply fits the NN to an MSE loss function without any modifications to the model architecture or loss formulation. This is particularly in contrast with the deep ensemble method, which uses the NLL as the loss function, and therefore obtains, in general, poorer RMSEs but better NLLs. 

\subsection{Chemistry applications}

To demonstrate the versatility of the LLPR approach, we considered its applicability to the task of learning the potential energies of molecules with neural network models. For this, we used the QM9 dataset~\cite{ramakrishnan2014quantum}, which contains approximately 130\,000 ground-state structures of small organic molecules, calculated by density functional theory~\cite{hohenberg1964inhomogeneous, kohn1965self}. As for the model, we adopted a Behler-Parrinello architecture~\cite{behler2007generalized} that takes SOAP atomic descriptors~\cite{bartok2013representing} as inputs, hereon referred to as a SOAP-BPNN model. More details are provided in~\ref{app:details}.

We defined the loss function for this exercise as
\begin{equation}
    \mathcal{L} = \sum_{i \in \mathcal{D}} \frac{(\tilde{y}_i - y_i)^2}{n_i},
\end{equation}
where $n_i$ is the number of atoms in chemical structure $i$. This loss is motivated by the assumption of additive atomic energy terms that are considered to be uncorrelated. This showcases the flexibility of the prediction rigidity framework, which can in principle be used with any loss function.

\begin{figure}[h]
  \centering
  \begin{subfigure}[b]{0.45\textwidth} %
    \includegraphics[width=\textwidth]{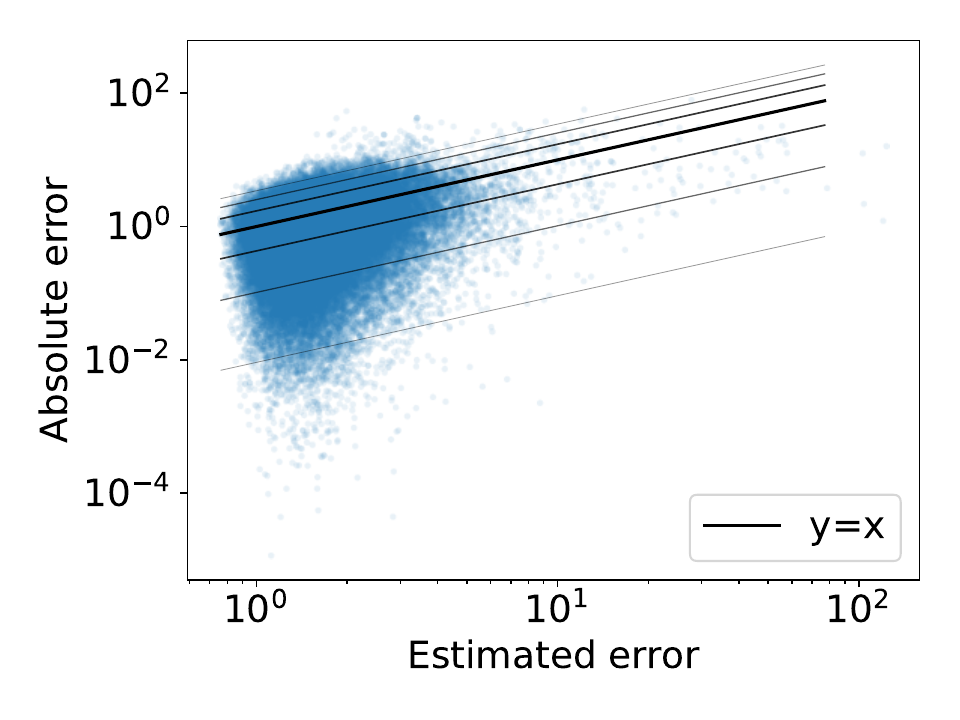}
  \end{subfigure}
  \begin{subfigure}[b]{0.45\textwidth} %
    \includegraphics[width=\textwidth]{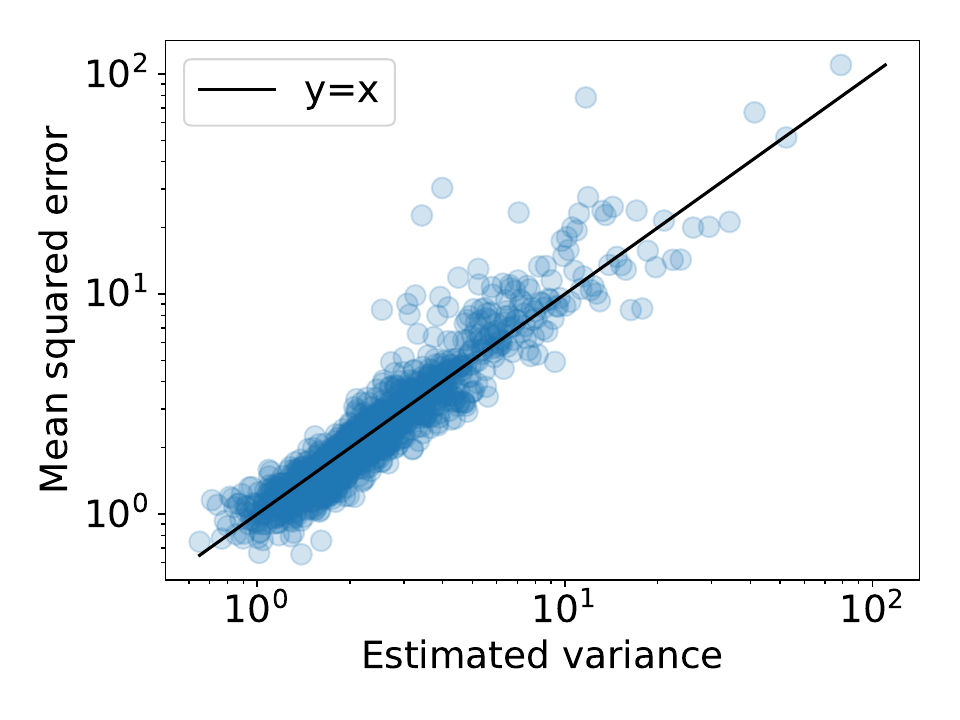}
  \end{subfigure}
  \caption{LLPR uncertainty estimates for a SOAP-BPNN model trained on the QM9 dataset. Left: parity plot of the estimated error vs absolute error on test samples. The thin black lines represent confidence intervals containing fractions of the probability distributions that are equal to those within one, two, and three standard deviations for a Gaussian distribution. Right: parity plot of the predicted variance vs mean squared error for the test samples, where each point is the average over a bin of 100 test set samples with similar estimated variances. More details on these plots can be found in~\ref{app:plot-details}.}
  \label{fig:qm9}
\end{figure}

Fig.~\ref{fig:qm9} shows that the LLPR formalism is able to recover excellent uncertainty estimates across several orders of magnitude in the error and for molecules of different sizes. The LLPR model correctly captures the neural network's confidence even on very uncertain outliers.

\subsection{Meteorology applications}

\begin{figure}[h]
  \centering
  \begin{subfigure}[b]{0.45\textwidth} %
    \includegraphics[width=\textwidth]{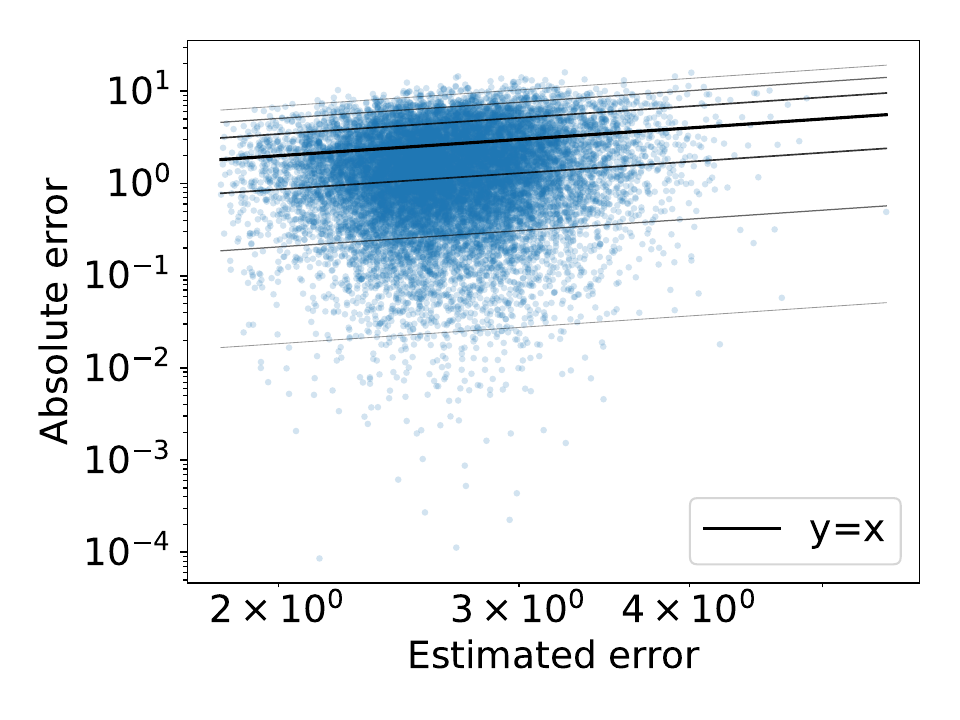}
  \end{subfigure}
  \begin{subfigure}[b]{0.45\textwidth} %
    \includegraphics[width=\textwidth]{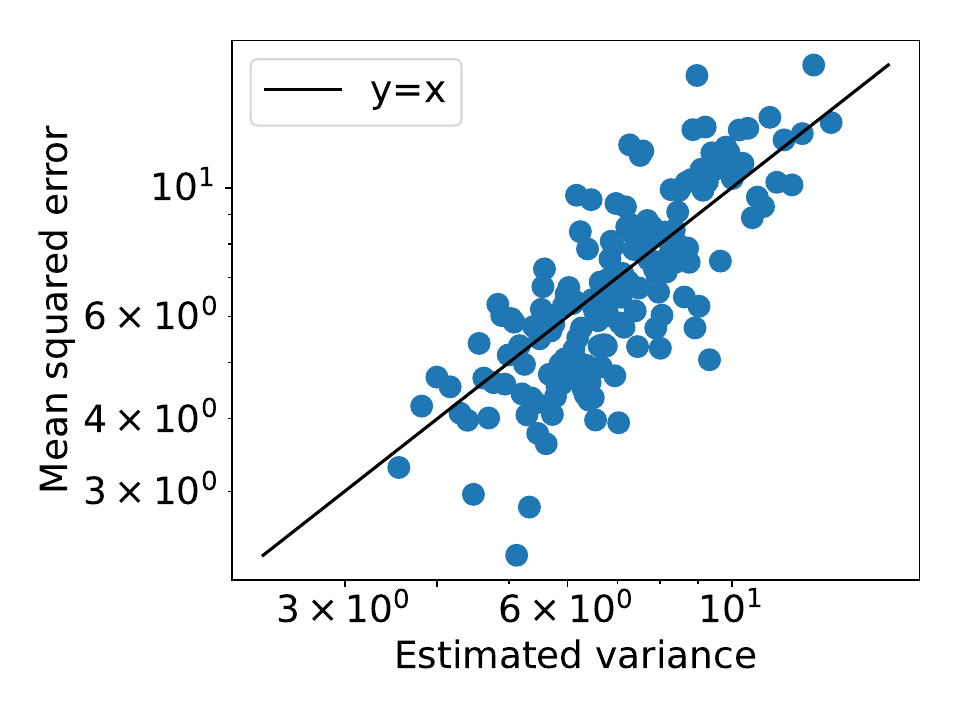}
  \end{subfigure}
  \caption{LLPR uncertainty predictions on the Australia weather dataset. Left: parity plot of the estimated error vs absolute error on test samples. The thin black lines represent confidence intervals containing fractions of the probability distributions that are equal to those within one, two, and three standard deviations for a Gaussian distribution. Right: parity plot of the predicted variance vs mean squared error for test samples. Each point represents the average of a bin of 200 test set samples with similar estimated variances. More details on the plots can be found in~\ref{app:plot-details}.}
  \label{fig:australia}
\end{figure}

We now move to a different application: that of weather forecasting. For this purpose, we take the ``Rain in Australia'' dataset \cite{australia_weather}, where we train a model to predict the next-day maximum temperatures starting from the available data on a given day. More details can be found in~\ref{app:details}. Figure~\ref{fig:australia} shows that the error estimates from the LLPR approach follow the expected distribution for this regression task as well.

\subsection{Out-of-domain detection}

\begin{figure}[h]
  \centering
  \begin{subfigure}[b]{0.45\textwidth} %
    \includegraphics[width=\textwidth]{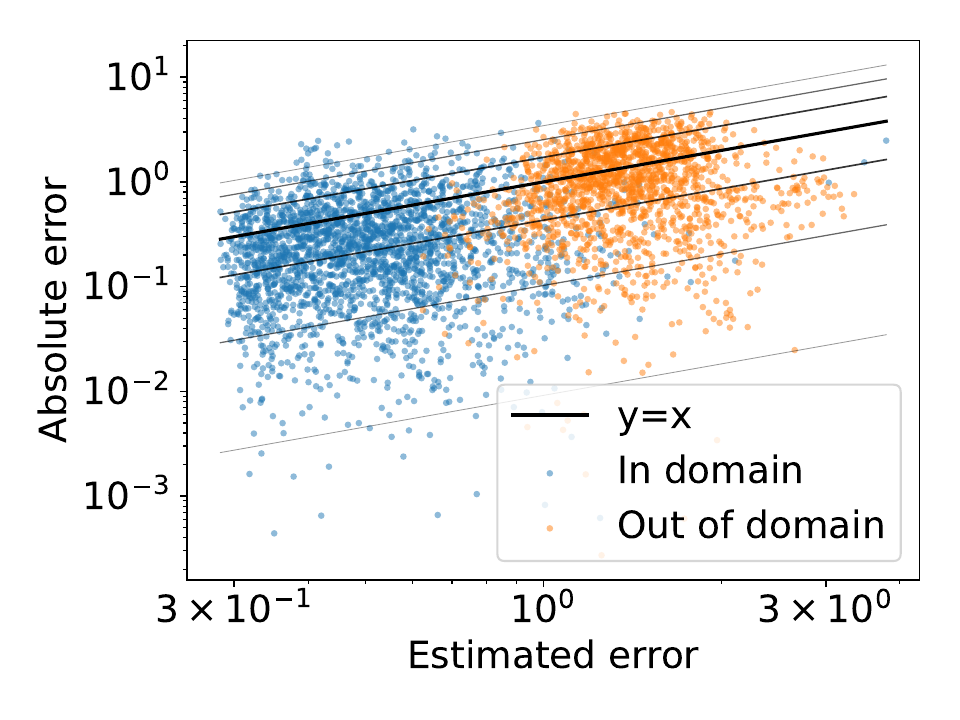}
  \end{subfigure}
  \begin{subfigure}[b]{0.45\textwidth} %
    \includegraphics[width=\textwidth]{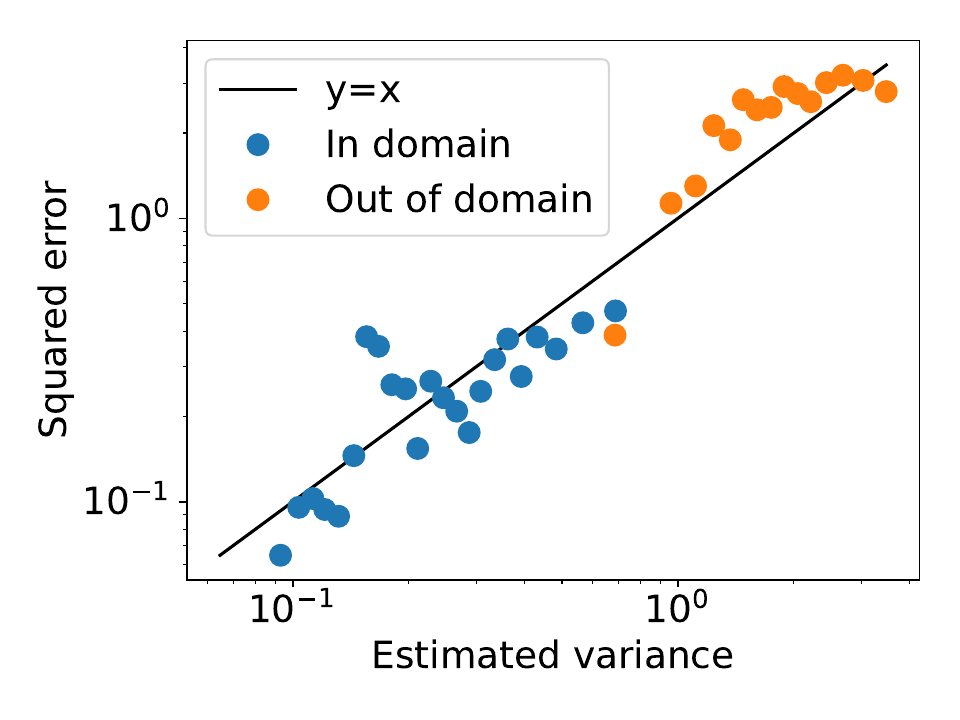}
  \end{subfigure}
  \caption{In-domain and out-of-domain uncertainty predictions on the California housing dataset. Left: parity plot of the estimated error vs absolute error on test samples. The thin black lines represent confidence intervals containing fractions of the probability distributions that are equal to those within one, two, and three standard deviations for a Gaussian distribution. Right: parity plot of the predicted variance vs mean squared error for test samples. Each point represents the average of a bin of 100 test set samples with similar predicted variance. More details on the plots can be found in~\ref{app:plot-details}.}
  \label{fig:ood_cali}
\end{figure}

In many applications, it is common for models to be queried outside of their training distribution. Therefore, one of the most important aspects of uncertainty quantification is the ability of the chosen confidence scheme to correctly detect highly uncertain out-of-domain samples. In order to test the LLPR approach in this context, we split the California housing dataset into two parts: one that only includes houses close to the ocean and one that only includes houses farther away from the ocean. The distinction is quite clear, as the distribution of this variable is bimodal (see~\ref{app:details} for details). We trained and validated the model on the far-from-the-ocean subset, then calculated the predictive uncertainties on both. In this scenario, the close-to-the-ocean subset is outside of the training and validation domain. Notice that the calibration described after Eq.~\eqref{eq:nngp-ntk-final-variance} was performed only on the far-from-the-ocean subset, as well. Figure \ref{fig:ood_cali} shows that LLPR-based uncertainty estimates can not only correctly ``flag'' the out-of-domain samples (since notably higher estimated errors are observed for the latter compared to the in-domain samples), but also provide highly accurate estimates in both scenarios.

\section{Discussion}

We have introduced the prediction rigidity as a theoretical framework to estimate predictive uncertainties in neural network models and other arbitrary regressors. Our method is based on the solution of a constrained optimization problem, which reflects the ``rigidity'' of the predictions of the model given its training set, and we have formalized a link between our framework and Bayesian inference and the Laplace approximation. Our method allows to obtain \textit{a posteriori} uncertainty estimates for any trained regressor, as demonstrated for polynomial, Gaussian, and neural network fits.

Our prediction-rigidity-based uncertainty estimation for neural networks relies on a last-layer approximation that was justified according to the theory of linearized neural network training. Besides its role in the present work, this framework helps rationalize why many works have observed last-layer approximations, which can might appear relatively crude at first sight, to provide satisfactory uncertainty estimates. We have shown that the predictive uncertainties obtained through last-layer prediction rigidities are competitive with established methods on standard benchmark datasets. Moreover, the proposed approach is effective in application scenarios, including the prediction of quantum mechanical properties of molecules and weather predictions, and it recovers high-quality uncertainty estimates even on out-of-distribution samples.

Besides providing good uncertainties, the proposed method is extremely convenient in practice. Indeed, it can be applied to arbitrary architectures, it is scalable to huge datasets and neural networks, it is easy to implement, and it can be applied to large pre-trained models. Moreover, it does not produce any significant overhead, neither during training nor during inference. As a result, last-layer prediction rigidities constitute a very promising method to estimate uncertainties in arbitrary neural networks with minimal human and computational effort. Finally, its potential for seamless integration with uncertainty propagation schemes, either analytically or numerically, makes this model ideal for any scientific or technological applications.

\section{Acknowledgements}

We thank Matthias Kellner for some insightful discussions. F.B. and M.C. acknowledge support from the NCCR MARVEL, funded by the Swiss National Science Foundation (grant number 182892). F.G., S.C., and M.C. acknowledge funding from the European Research Council (ERC) under the European Union’s Horizon 2020 research and innovation programme (grant no. 101001890-FIAMMA). S.C. and M.C. acknowledge the financial support by the Swiss National Science Foundation (Project 200020\_214879).

\section{Data availability statement}

The supporting data for this work, including datasets, are available on Zenodo~\cite{zenodo} at~\href{https://zenodo.org/records/10775137}{https://zenodo.org/records/10775137}.

\bibliographystyle{unsrt}

\appendix
\section{Prediction rigidities for linear models and Gaussian process regression}\label{app:linear-gpr}

\subsection{Linear models}

In a linear model, if $\mathbf{y}$ is the vector collecting all the targets in the training set and $\mathbf{X}$ is the feature matrix whose rows corresponds to different training samples and columns correspond to different features, we have
\begin{equation}
    \mathcal{L}(\mathbf{w}) = (\mathbf{y} - \mathbf{X} \mathbf{w})^\top (\mathbf{y} - \mathbf{X} \mathbf{w}) + \mathbf{w}^\top \mathbf{\Sigma} \mathbf{w},
\end{equation}
where $\mathbf{\Sigma}$ is a regularization matrix, and
\begin{equation}
    y_\star = \mathbf{w}^\top \mathbf{x}_\star. 
\end{equation}
Therefore, the prediction rigidity framework affords
\begin{equation}
    \sigma_\star^2 \propto R_\star^{-1} = \frac{\partial \tilde{y}_\star}{\partial\mathbf{w}} \Big|^\top_{\mathbf{w}_o} \mathbf{H}_o^{-1} \frac{\partial \tilde{y}_\star}{\partial \mathbf{w}} \Big|_{\mathbf{w}_o} = \mathbf{x}_\star^\top (\mathbf{X}^\top \mathbf{X} + \mathbf{\Sigma})^{-1} \mathbf{x}_\star,
\end{equation}
which corresponds to the well-known uncertainty formula for linear regression.

\subsection{Gaussian process regression}

We will now consider the case of Gaussian process regression in the subset of regressors formalism, where the loss function is
\begin{equation}
    \mathcal{L}(\mathbf{w}) = (\mathbf{y} - \mathbf{K}_{nm} \mathbf{w})^\top (\mathbf{y} - \mathbf{K}_{nm} \mathbf{w}) + \sigma^2 \mathbf{w}^\top \mathbf{K}_{mm} \mathbf{w},
\end{equation}
and predictions are given by
\begin{equation}
    y_\star = \mathbf{w}^\top \mathbf{k}_\star. 
\end{equation}
As a result, the prediction rigidity framework predicts
\begin{equation}
    \sigma_\star^2 \propto R_\star^{-1} = \frac{\partial \tilde{y}_\star}{\partial\mathbf{w}} \Big|^\top_{\mathbf{w}_o} \mathbf{H}_o^{-1} \frac{\partial \tilde{y}_\star}{\partial \mathbf{w}} \Big|_{\mathbf{w}_o} = \mathbf{k}_\star^\top (\mathbf{K}_{nm}^\top \mathbf{K}_{nm} + \sigma^2\mathbf{K}_{mm})^{-1} \mathbf{k}_\star,
\end{equation}
which indeed corresponds to the formula for uncertainty predictions in the subset of regressors formulation of Gaussian process regression \cite{rasmussen2006gaussian}.

\section{Power series expansion of the NTK}\label{app:NTK linearization}

For a $L$-layer NN (with NTK-parametrization):
\begin{equation}
    \tilde{y}_i \equiv 
    \mathbf{W}^{(L)}\frac{1}{\sqrt{N_L}} 
    \phi
    \left( \mathbf{W}^{(L-1)}\frac{1}{\sqrt{N_{L-1}}} 
    \phi\Big(
    \ldots \, \mathbf{W}^{(1)}\frac{1}{\sqrt{N_1}} \phi(\mathbf{W}^{(0)} \mathbf{x}_i) \ldots
    \Big) 
    \right) 
\end{equation}
where $\mathbf{W}^{(l)}$ is an $ N_{l+1} \times N_{l}$ matrix of weights whose entries are sampled from $\mathcal{N}(0,1)$ [we set $\sigma_w=1$ for simplicity], $N_0 = \mathrm{len}(\mathbf{x}_i)$, $N_{L+1} = 1$, and $N_{1\leq l\leq L} \to \infty$ for infinite-width NN, the NTK between two inputs $\mathbf{x}_i$ and $\mathbf{x}_j$ can be obtained according to the following recursion formulas \cite{jacot2018neural}:
\begin{equation}\label{eq:NTK-recursive}
\begin{split}
    &K_\nngp^{(0)} (\mathbf{x}_i,\mathbf{x}_j) = K_\ntk^{(0)} = \mathbf{x}_i^\top \mathbf{x}_j \\
    &K_\nngp^{(l)} (\mathbf{x}_i,\mathbf{x}_j) = \check{\phi}\left(K_\nngp^{(l-1)}(\mathbf{x}_i,\mathbf{x}_j)\right) \\
    &K_\ntk^{(l)} (\mathbf{x}_i,\mathbf{x}_j) = K_\nngp^{(l)}(\mathbf{x}_i,\mathbf{x}_j) + K_\ntk^{(l-1)} (\mathbf{x}_i,\mathbf{x}_j) \, \check{\phi}' \left(K_\nngp^{(l-1)}(\mathbf{x}_i,\mathbf{x}_j)\right).
\end{split}
\end{equation}
Here, $\check{\phi}$ is the \textit{dual activation} for the activation function $\phi$, defined by~\cite{daniely2016toward}:
\begin{equation}\label{eq:dual_phi}
    \check{\phi}(\xi) \equiv C \iint \phi(u) \phi(v) \exp
    \left[-\frac{1}{2}{
    \begin{pmatrix}
        u & v
    \end{pmatrix}
    \begin{pmatrix}
        1   & \xi \\
        \xi & 1
    \end{pmatrix}
    \begin{pmatrix}
        u \\ 
        v
    \end{pmatrix}
    } 
    \right]
    \mathrm{d}u \, \mathrm{d}v
\end{equation}
and $C$ is the normalization constant such that $\check{\phi}(1) = 1$.
If $\phi(z)$ is any function linearizable in $z=0$ (e.g.~$\tanh$, $\mathrm{erf}$, $\mathrm{SiLU}$, $\mathrm{GELU}$), we can  Taylor expand its dual:
\begin{equation}
    \check{\phi}(\xi) = \sum_{n=0}^{\infty} \frac{1}{n!} \check{\phi}^{(n)}(0) \, \xi^n =
    \check{\phi}(0) 
    + \check{\phi}'(0) \, \xi 
    + \frac{1}{2!} \check{\phi}''(0) \, \xi^2
    + \frac{1}{3!} \check{\phi}'''(0) \, \xi^3
    + \ldots
\end{equation}
Since the dual commutes with differentiation (see Suppl.~Sec.~C of Ref.~\cite{daniely2016toward}), i.e.~$(\check{\phi})' = \check{(\phi')}$, we can study the integral 
\begin{equation}
    \check{\phi}^{(n)}(0) = C \left[\int \phi^{(n)}(u) e^{-\frac{1}{2}u^2} \mathrm{d}u \right]^2 
\end{equation}
when $\phi(u)$ is Taylor expanded around $z=0$ (for the special case of ReLU, we redirect the reader to Ref.~\cite{kristiadi2020being}). In particular, whenever $\phi(u)$ is odd (exact for $\tanh(u)$, $\mathrm{erf}(u)$, and still a good approximation for $\mathrm{SiLU}(u)$, $\mathrm{GELU}(u)$ in the range $u\in [-1,1]$ where $\phi(u)$ is not suppressed by the factor $e^{-\frac{1}{2}u^2}$ under integration), all the even derivatives $\check{\phi}^{(n)}(0)$ vanish, which leads to
\begin{equation}
\begin{split}
    &\check{\phi}(\xi) = \check{\phi}'(0) \, \xi + \frac{1}{3!}\, \check{\phi}'''(0) \, \xi^3 + \mathcal{O}(\xi^5) \\
    &\check{\phi}'(\xi) = \check{\phi}'(0) + \frac{1}{2}\, \check{\phi}'''(0) \, \xi^2 + \mathcal{O}(\xi^4).
\end{split}
\end{equation}
If we replace these expressions into Eq.~\eqref{eq:NTK-recursive}, we obtain:
\begin{equation}\label{eq:NTK-recursive-Taylor}
\begin{split}
    K_\nngp^{(0)} (\mathbf{x}_i,\mathbf{x}_j) &= \mathbf{x}_i^\top \mathbf{x}_j = K_\ntk^{(0)} \\
    K_\nngp^{(1)} (\mathbf{x}_i,\mathbf{x}_j) &= \check{\phi}\left(\mathbf{x}_i^\top \mathbf{x}_j\right) = \check{\phi}'(0) \mathbf{x}_i^\top \mathbf{x}_j + \mathcal{O}[(\mathbf{x}_i^\top \mathbf{x}_j)^3]\\
    K_\ntk^{(1)} (\mathbf{x}_i,\mathbf{x}_j) &= \check{\phi}'(0) \,\mathbf{x}_i^\top \mathbf{x}_j + \mathbf{x}_i^\top \mathbf{x}_j \, \check{\phi}'(0) + \mathcal{O}[(\mathbf{x}_i^\top \mathbf{x}_j)^3] = 2 \check{\phi}'(0) \mathbf{x}_i^\top \mathbf{x}_j + \mathcal{O}[(\mathbf{x}_i^\top \mathbf{x}_j)^3] \\
    &\ldots \\
    K_\nngp^{(l)} (\mathbf{x}_i,\mathbf{x}_j) 
    &= [\check{\phi}'(0)]^l \, \mathbf{x}_i^\top \mathbf{x}_j + \mathcal{O}[(\mathbf{x}_i^\top \mathbf{x}_j)^3]\\
    K_\ntk^{(l)} (\mathbf{x}_i,\mathbf{x}_j) &= (l+1) [\check{\phi}'(0)]^l \mathbf{x}_i^\top \mathbf{x}_j + \mathcal{O}[(\mathbf{x}_i^\top \mathbf{x}_j)^3] \\
    &\ldots
\end{split}
\end{equation}
In particular, for the last layer, at initialization, we have, up to $\mathcal{O}[(\mathbf{x}_i^\top \mathbf{x}_j)^3]$
\begin{equation}
    K_\ntk^{(L)} (\mathbf{x}_i,\mathbf{x}_j)   \approx (L+1) \, K_\nngp^{(L)} (\mathbf{x}_i,\mathbf{x}_j) 
    \approx (L+1)\, \mathbf{f}_i^\top \mathbf{f}_j.
\end{equation}

One may then wonder why not using directly the initial features to construct the NTK even at the last layer, since $K_\ntk^{(L)}(\mathbf{x}_i, \mathbf{x}_j) \propto \mathbf{x}_i^\top \mathbf{x}_j$, up to $\mathcal{O}[(\mathbf{x}_i^\top \mathbf{x}_j)^3]$. Indeed it can be shown by induction that the $\mathcal{O}[(\mathbf{x}_i^\top \mathbf{x}_j)^3]$ correction is larger for the latter case. In particular, for $l>0$ [to lighten the notation we set $a = \check{\phi}'(0)$ and $b = \check{\phi}'''(0)$]:
\begin{equation}
\begin{split}
    \Xi^{(l)}(\mathbf{x}_i,\mathbf{x}_j) &\equiv K_\ntk^{(l)}(\mathbf{x}_i,\mathbf{x}_j) - (l+1) \, a^l \, \mathbf{x}_i^\top \mathbf{x}_j \\
    &= \left[\frac{1}{6} a^{l -1} \sum_{m=1}^{l} a^{2 (m -1)} (2m +l +1) 
    \right] 
    b\,(\mathbf{x}_i^\top \mathbf{x}_j)^3 + \mathcal{O}[(\mathbf{x}_i^\top \mathbf{x}_j)^5],
\end{split}
\end{equation}
while
\begin{equation}
\begin{split}
    \Delta^{(l)}(\mathbf{x}_i,\mathbf{x}_j) &\equiv K_\ntk^{(l)}(\mathbf{x}_i,\mathbf{x}_j) - (l+1)K_\nngp^{(l)}(\mathbf{x}_i,\mathbf{x}_j)  \\
    &= \left[\frac{1}{3} a^{l -1} \sum _{m =1}^{l} m \, a^{2 (m -1)}
    \right] 
    b \,(\mathbf{x}_i^\top \mathbf{x}_j)^3 + \mathcal{O}[(\mathbf{x}_i^\top \mathbf{x}_j)^5].
\end{split}
\end{equation}
In the Zenodo repository associated to our work, we provide a Mathematica~\cite{Mathematica} notebook addressing the set of calculations needed for the proof by induction.
Assuming $a>0$, the difference between the norm of these contributions is
\begin{equation}
    \left|\Xi^{(l)}(\mathbf{x}_i,\mathbf{x}_j)\right| - \left|\Delta^{(l)}(\mathbf{x}_i,\mathbf{x}_j)\right| = 
    \left[\frac{1}{6} a^{l -1} \, (l+1) \sum_{m=1}^{l} a^{2 (m -1)}  
    \right]
    \left|
    b \,(\mathbf{x}_i^\top \mathbf{x}_j)^3
    \right|
    + \mathcal{O}[(\mathbf{x}_i^\top \mathbf{x}_j)^5].
\end{equation}
The term between square brackets is positive:
\begin{equation}
    \frac{1}{6} a^{l -1} (l+1)  \sum_{m=1}^{l} a^{2 (m -1)} 
    = \frac{1}{6} a^{l -1} (l+1) \frac{\left(a^{2 l}-1\right)}{ \left(a^2-1\right)} > 0,
\end{equation}
which implies that $K_\ntk^{(L)}  \approx (L+1) \,  K_\nngp^{(L)} (\mathbf{x}_i,\mathbf{x}_j) \approx (L+1)\, \mathbf{f}_i^\top \mathbf{f}_j $ is a better approximation than $K_\ntk^{(L)}(\mathbf{x}_i, \mathbf{x}_j) \approx (L+1) \,[\check{\phi}'(0)]^L \, \mathbf{x}_i^\top \mathbf{x}_j$.
As stressed, to obtain a good estimator for $K_\nngp^{(L)} (\mathbf{x}_i,\mathbf{x}_j)$, one should in principle employ the empirical $\mathbf{f}_i$ at initialization. Nonetheless, $\mathbf{F} \mathbf{F}^\top$ does not change much during training, as we have explicitly verified in the cases analyzed in this work. In general, we found $\mathbf{F} \mathbf{F}^\top$ to vary less than the full NTK, i.e., $K_\ntk^{(L)}$, during learning. This is reasonable because $\mathbf{f}_i \equiv \partial \tilde{y}_i / \partial \mathbf{W}^{(L)}$ are independent of the weights $\mathbf{W}^{(L)}$, which are the weights reported to change more during training (see, e.g.,~\cite{jacot2018neural} and its reviews). In line with the last-layer Laplace approximation, this observation concludes our reasoning to motivate Eq.~\eqref{eq:NTK_multiple_FF^T}. Whenever $\sigma_w \neq 1$ is introduced, or other parametrizations are used to sample the weights at initialization \cite{sohl2020infinite}, the prefactor is modified from $L+1$ [this is indeed why we kept a generic constant $c$ in Eq.~\eqref{eq:NTK_multiple_FF^T}] but it remains a constant independent of the data sample. 

\section{The effect of biases}\label{app:biases}
In the presence of a bias in the final layer of a neural network, the NNGP can be written, similarly to Eq.~\eqref{eq:nngp-approx}, as
\begin{equation}
    \mathbf{K}_\nngp(\mathcal{D}, \mathcal{D}) \approx \sigma_b^2 \mathbf{J} + \sigma_w^2 \, \mathbf{F} \mathbf{F}^\top,
\end{equation}
where $\mathbf{J}$ is a matrix filled with ones, and the NTK can be written as
\begin{equation}\label{eq:ntk-bias}
    K_\ntk = \mathbf{J} + \mathbf{F} \mathbf{F}^\top.
\end{equation}
In practice, the standard deviation of the biases might match that of the weights (this is, for example, the default behavior of linear layers in PyTorch~\cite{paszke2019pytorch}). In that case, the two kernels are proportional to one another and an analogous derivation to the one in the main text holds. We accounted for biases in this way in all results.

Another possibility is that, as the number of hidden features increases, the influence of the bias vanishes in practice, so that the formulas in the main text approximately hold. Although we did not neglect the biases in our final results, their omission yields very similar uncertainty estimates when compared to their explicit inclusion (Eq.~\eqref{eq:ntk-bias}).

\section{The role of the regularizer}\label{app:regularizer}

In Eq.~\eqref{eq:woodbury-trick}, we introduced a regularizing term $\varsigma^2$ without explaining its relevance. This term serves mainly two purposes:
\begin{itemize}
    \item Numerical stability: although the Hessian in the last-layer PR is positive semi-definite, this does not guarantee its invertibility. A simple case where the Hessian matrix is not invertible is that of a training set which contains the same training sample multiple times. Similarly, the inversion could become numerically unstable if the training dataset contains samples that are very similar to one another. For this reason, it is helpful to include a small regularization parameter.
    \item Regularization in NNs: $\varsigma^2$ takes into account the various regularization techniques that are used, explicitly or implicitly, to train neural networks. These can include early stopping, weight decay, and others. In our experiments, we found that the optimal value of $\varsigma^2$ was close to zero if no weight decay was used, but it was relatively large instead if weight decay was applied. On the other hand, we found the correlation between early stopping and the optimal value of the regularizing term to be less consistent.
\end{itemize}

\section{Behavior as a function of the number of neurons}\label{app:width}

\begin{figure}[h]
\centering
\begin{subfigure}[b]{0.49\linewidth}
\includegraphics[width=\linewidth]{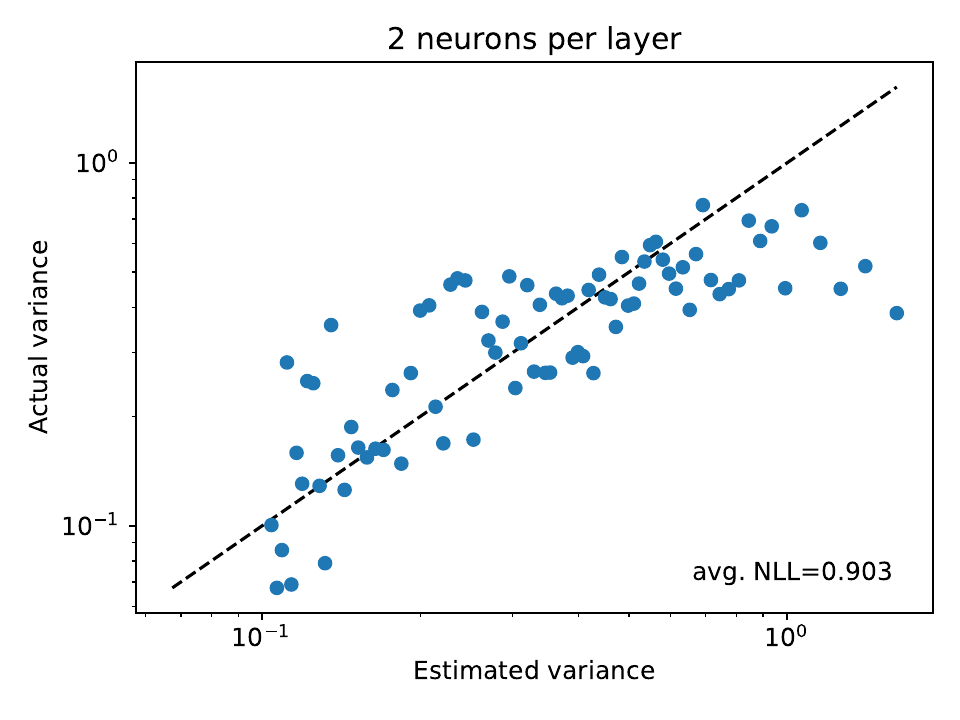}
\end{subfigure}
\begin{subfigure}[b]{0.49\linewidth}
\includegraphics[width=\linewidth]{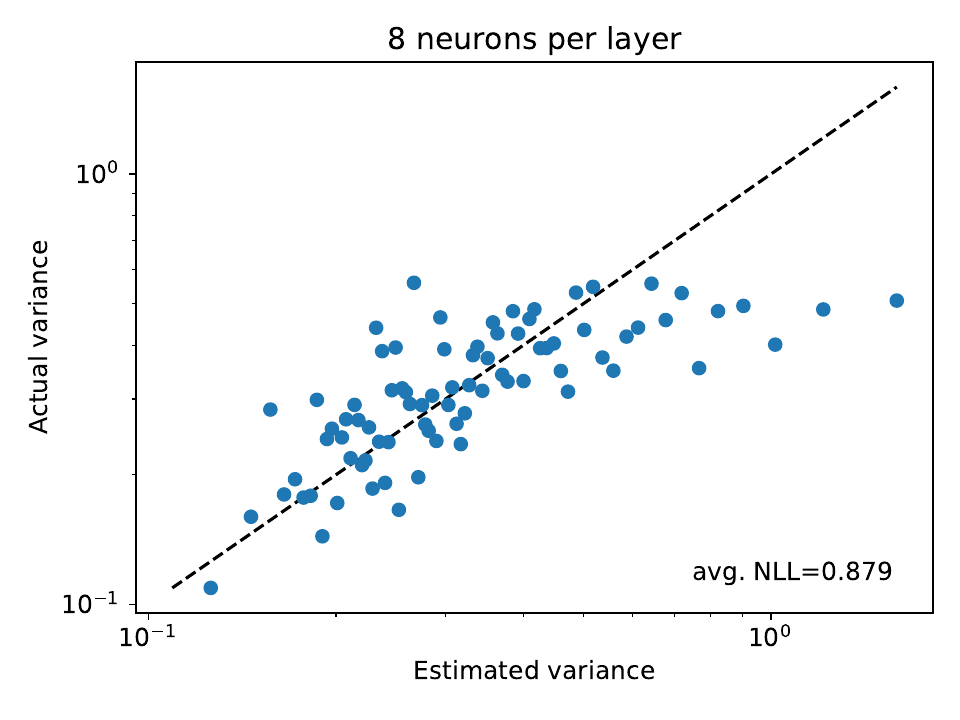}
\end{subfigure}
\begin{subfigure}[b]{0.49\linewidth}
\includegraphics[width=\linewidth]{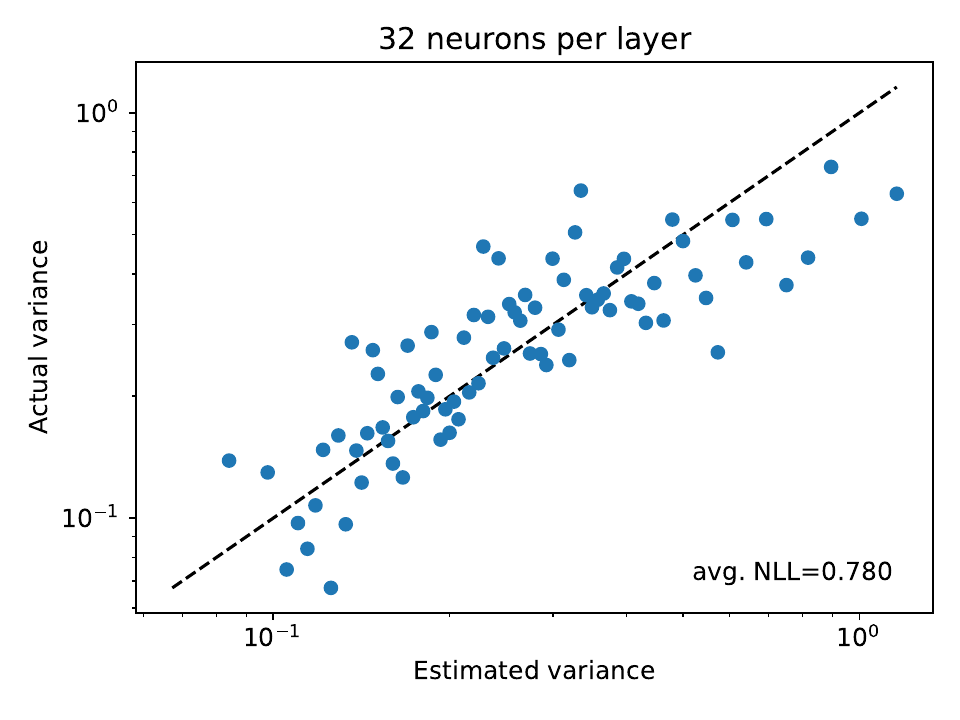}
\end{subfigure}
\begin{subfigure}[b]{0.49\linewidth}
\includegraphics[width=\linewidth]{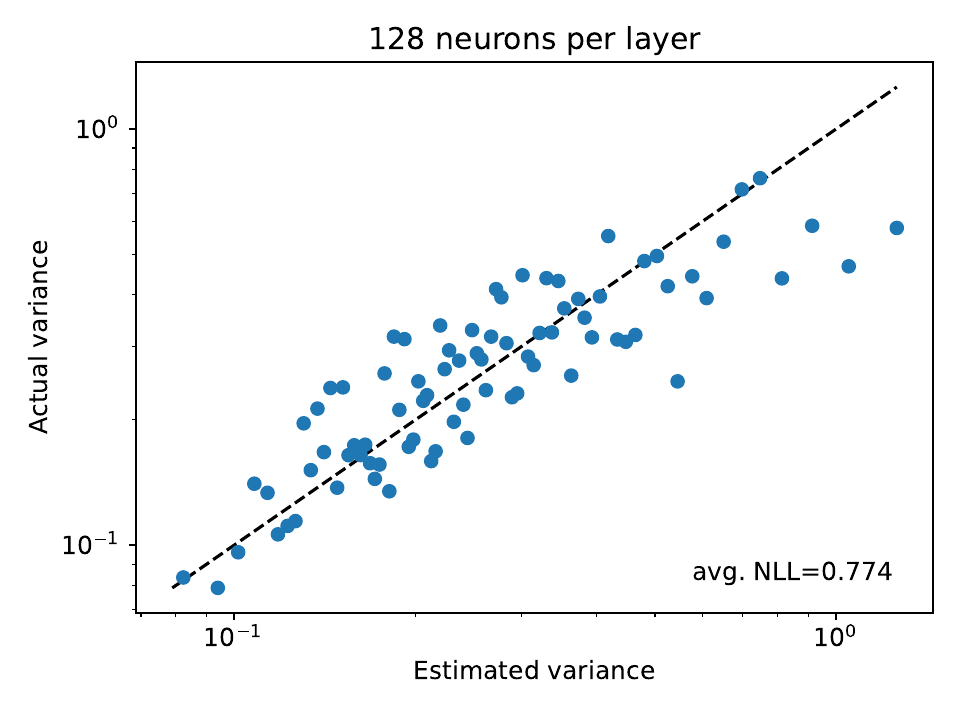}
\end{subfigure}

\caption{Quality of the LLPR uncertainty estimates as a function of the number of neurons per layer. Each point corresponds to the average of 100 test samples.}
\label{fig:n_neurons}
\end{figure}

According to the theoretical justification of the last-layer PR in Section~\ref{sec:nn-application}, last-layer prediction rigidities are expected to become more reliable as the width of the neural network increases. In Figure \ref{fig:n_neurons}, we explore the quality of predicted uncertainties on the California housing dataset as a function of the number of neurons.

We split the California housing dataset into 20\% train / 20\% validation, 60\% test splits. The architecture consisted of a multi-layer perceptron with 2 layers and SiLU activation functions. For this specific dataset, we disabled biases to isolate the effect of the varying number of weights.

Indeed, figure \ref{fig:n_neurons} shows that the quality of uncertainty predictions of the last-layer prediction rigidity framework improves as the number of neurons per layer grows. For the datasets which were explored in this work, we found 50 or more neurons to consistently afford high-quality uncertainty estimates, although in many cases we obtained good results with fewer neurons.

\section{Dataset and fit details}\label{app:details}

\subsection{General remarks}

The two optimizable parameters in the last-layer PR formula were optimized by grid search on the same validation set that was used to monitor the training process. The target of this optimization was the sum of squared residuals of averages of bins of 100 from the $y=x$ line (on an expected variance vs actual MSE plot), except for the PBP benchmark, where the optimization target was the validation NLL. Furthermore, unless otherwise specified: 
\begin{itemize}
    \item The SiLU activation function was employed.
    \item The neural network fitting scheme consisted of training for 400 epochs with the AdamW optimizer and learning rate reduction by a factor of 10 upon stagnation of the validation loss for 100 epochs. At the end of this procedure, the parameters that afforded the best validation loss were chosen.
\end{itemize}

\subsection{Simple fits}

For this experiment, we take the $y = \cos^2 x$ function, we add random Gaussian noise with standard deviation equal to 0.01, and the training set corresponds to x = -0.8, -0.75, 0.0, 0.05, 0.07, 0.7, 0.73. Polynomial fitting and evaluation were performed with the JAX~\cite{bradbury2018jax} version of the NumPy~\cite{harris2020array} \texttt{polyfit} and \texttt{polyeval} functions. The Gaussian fit was obtained as a linear combination of two Gaussian functions each with optimizable prefactor, mean and variance. The fit was performed with the L-BFGS algorithm. The neural network fit was the result of training a multi-layer perceptron with 2 hidden layers and 32 neurons per hidden layer. The uncertainty estimates for this neural network were obtained with the last-layer PR approximation.

\subsection{PBP benchmark}

We took the datasets from the MC Dropout \cite{gal2016dropout} official repository, excluding the Boston housing dataset due to ethical concerns. We found that models in the literature are trained according to different protocols. In our experiments, we decided to be consistent with the one used by MC dropout \cite{gal2016dropout}, although this makes our protocol inconsistent with the training protocol followed in the Deep Ensemble paper~\cite{lakshminarayanan2017simple}. However, we experimented with our own implementation of Deep Ensembles and trained them according to the MC Dropout protocol. We did not find significant differences with the literature results.

Furthermore, instead of the canonical 90\% train / 10\% validation split that was employed in previous works, we employed a 80\% train / 10\% validation / 10\% test split, which allowed the validation set to be used to calibrate the LLPR uncertainty estimates.

\subsection{Chemistry}

After removing the known inconsistent structures from the QM9 dataset, we split 10,000 random training and 10,000 random validation structures, while the rest was kept for testing.
We used the SOAP molecular descriptors~\cite{bartok2013representing} computed with the \texttt{rascaline-torch} library. These descriptors were used as the inputs to a Behler-Parrinello neural network architecture~\cite{behler2007generalized} with 3 layers of 64 neurons per chemical species. Unlike the other models trained in this work, this was trained (and evaluated) in 64-bit floating-point precision.

\subsection{Meteorology}

We employed the Rain in Australia dataset~\cite{australia_weather}, and we pre-processed it in the following way:
\begin{itemize}
    \item We set the target to be the maximum temperature on the following day.
    \item We did not use the ``Date'', ``RainToday'' and ``RainTomorrow'' variables.
    \item We dropped any entries with any unavaliable remaining variables.
    \item We converted each wind direction variable (given in the original dataset as one of 16 wind directions: N, NNE, NE, ...) into two continuous variables, given by their corresponding x and y values on the unit circle.
\end{itemize}
This affords a dataset with 26 continuous variables and one discrete variable (the location of the weather station). We split the resulting dataset into a 40\% train, 30\% validation, 30\% test split. Training was performed with a neural network of 2 layers and 256 neurons per layer. The input features for this neural network consisted of the 26 continuous variable and a 256-dimensional learnable embedding of the location of the weather station.

\subsection{Out-of-distribution California housing}

From a simple analysis of the input variables in the California housing dataset, it can be noticed that the ``ocean distance'' input follows a bimodal distribution to a good approximation.

\begin{figure}[h]
  \centering
  \includegraphics[width=0.6\textwidth]{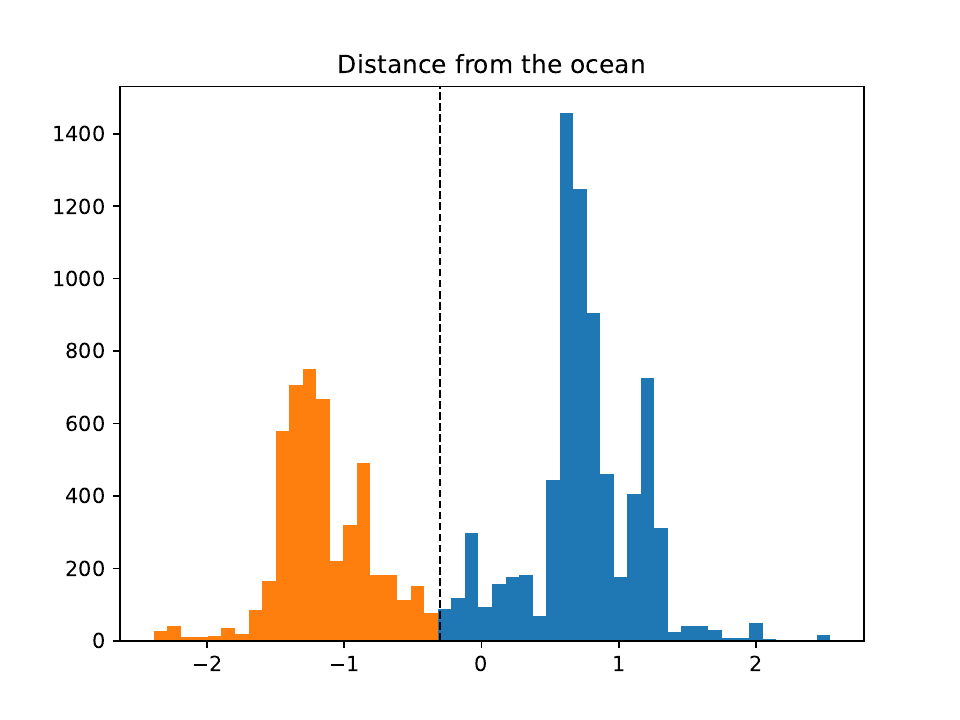}
  \caption{Distribution of the normalized eighth feature in the California housing dataset (ocean distance). We selected the in-domain samples to be those whose eighth feature is 
 greater than -0.3, while we considered all other samples to be out of domain.}
  \label{fig:ood_cali_histogram}
\end{figure}

Given this observation, and after splitting the dataset into 60\% train / 20\% validation / 20\% test, we discarded all train and validation samples whose centered and normalized distance from the ocean was less than -0.3 (see figure \ref{fig:ood_cali_histogram}).
In the same way, we divided the test set into the in-distribution and out-of-distribution subsets whose uncertainties we plotted in Fig.~\ref{fig:ood_cali} after training. Training was performed with a neural network with 2 hidden layers and 128 neurons per layer.

\subsection{Plot details}\label{app:plot-details}

Throughout this work, two types of parity plots are provided. In order to fully understand them, it is useful to consider the expected theoretical distribution of the errors of a model at a given predicted uncertainty. If the model predicts uncertainties correctly, the distribution of the errors will be approximately Gaussian, with zero mean and standard deviation $\sigma$ given by the error estimate of the model. 

Therefore, the resulting \textit{absolute} error distribution will be the ``folding'' of the original Gaussian distribution with zero mean that occurs when negative inputs are turned into positive ones. This would be the ideal distribution of points in a single vertical slice of our non-binned plots. However, the logarithmic scale turns this distribution into a new distribution whose probability distribution function is
\begin{equation}
    P(x) \propto x \, e^{-\frac{x^2}{2\sigma^2}}.
\end{equation}
The mode of this distribution is at $x=\sigma$ (this can be seen easily by taking the first derivative and setting it to zero), and this (asymmetric) distribution is the expected distribution of each vertical slice of our non-binned plots. The confidence lines we show in such figures are chosen so as to contain a fraction of the total probability equivalent to that contained within one, two and three standard deviations of a Gaussian distribution, while simultaneously imposing that any two corresponding confidence lines (above and below the mode) evaluate to the same probability distribution function value. The latter additional constraint is necessary to define the position of the confidence lines (note that the distribution is asymmetric).

Our binned parity plots instead contain the estimated variance vs the mean squared error within a bin, where bins are created according to the estimated variance of the model. If not for the binning, almost nothing would change, since raising both axes to the power of two only shifts the scale of the log-log plot. However, the binning is used for the purpose of highlighting that the MSE distribution indeed has the correct mode on a logarithmic scale (in this case, $\sigma^2$, where $\sigma^2$ is the predicted variance of the model).

\end{document}